# scDrugMap: Benchmarking Large Foundation Models for Drug Response Prediction

Qing Wang[1,†], Yining Pan[1,†], Minghao Zhou[1,†], Zijia Tang[2], Yanfei Wang[1], Guangyu Wang[3,4], Qianqian Song[1,*]

[1]Department of Health Outcomes and Biomedical Informatics, University of Florida, Gainesville, FL 32611, USA

[2]Trinity College, Duke University, Durham, NC, USA

[3]Center for Bioinformatics and Computational Biology, Houston Methodist Research Institute, Houston, TX, USA

[4]Department of Cardiothoracic Surgery, Weill Cornell Medicine, Cornell University, New York, NY, USA

† The authors contributed equally to this work.

*Correspondence: qsong1@ufl.edu

## ABSTRACT

Drug resistance remains a significant barrier to improving the effectiveness of cancer therapies. To better understand the biological mechanisms driving resistance, single-cell profiling has emerged as a powerful tool for characterizing cellular heterogeneity. Recent advancements in large-scale foundation models have demonstrated potential in enhancing single-cell analysis, yet their performance in drug response prediction remains underexplored.

In this study, we developed scDrugMap, an integrated framework for drug response prediction that features both a Python command-line tool and an interactive web server. scDrugMap supports the evaluation of a wide range of foundation models, including eight single-cell foundation models and two large language models (LLMs), using large-scale single-cell datasets across diverse tissue types, cancer types, and treatment regimens. The framework incorporates a curated data resource consisting of a primary collection of 326,751 cells from 36 datasets across 23 studies, and a validation collection of 18,856 cells from 17 datasets across 6 studies. Using scDrugMap, we conducted comprehensive benchmarking under two evaluation scenarios: pooled-data evaluation and cross-data evaluation. In both settings, we implemented two model training strategies—layer freezing and fine-tuning using Low-Rank Adaptation (LoRA) of foundation models.

In the pooled-data evaluation, scFoundation outperformed all others, while most models achieved competitive performance. Specifically, scFoundation achieved the highest mean F1 scores of 0.971 and 0.947 using layer-freezing and fine-tuning, outperforming the lowest-performing model by 54% and 57%, respectively. In the cross-data evaluation, UCE achieved the highest performance (mean F1 score: 0.774) after fine-tuning on tumor tissue, while scGPT demonstrated superior performance (mean F1 score: 0.858) in a zero-shot learning setting. Together, this study presents the first comprehensive benchmarking of large-scale foundation models for drug response prediction in single-cell data and introduces a user-friendly, flexible platform to support drug discovery and translational research.

# INTRODUCTION

Drug resistance continues to be a significant challenge in disease treatment, particularly in cancer therapy[1]. An analysis of 85 cancer drugs approved by the U.S. Food and Drug Administration (FDA) revealed a median response rate (RR) of 41%, with nearly half exhibiting an RR below 40%, and two-thirds showing a complete RR of less than 10%[2]. Even in biomarker-based personalized cancer treatments, response rates remain modest—30.6% compared to just 4.9% for non-biomarker-based therapies[3]. Moreover, drug resistance is closely associated with patient survival rates. For example, resistance to KRAS inhibitors, a standard treatment for non-small cell lung cancer (NSCLC), is associated with a median survival of only 6.3 months[4]. Similarly, in glioblastoma, resistance to chemotherapy and radiotherapy limits median survival post-diagnosis to just 12-15 months, with less than 10% of patients surviving longer than five years[5]. These findings underline the pressing need for further investigation and advanced approaches to better elucidate drug resistance mechanisms and improve treatment efficacy.

Over the past decade, single-cell RNA sequencing (scRNA-seq) technology has revolutionized our understanding of cellular and molecular heterogeneity, enabling high-resolution analysis of drug responses across diverse cell types and subtypes[6-13]. For instance, scRNA-seq analysis of NSCLC tumor tissue identified elevated expression of estrogen metabolism enzymes and serum estradiol in cancer cells exhibiting poor pathological responses to PD-1 blockade immunotherapy. Additionally, aged CCL3+ neutrophils interacting with tumor-associated macrophages were identified as potential biomarkers of poor therapy response[13]. In basal cell carcinoma (BCC), scRNA-seq revealed that novel T cell recruitment, rather than the reinvigoration of pre-existing tumor-infiltrating lymphocytes, was pivotal for effective PD-1 blockade therapy[11]. Similarly, in inflammatory breast cancer (IBC), single-cell profiling uncovered a shift in cell state from luminal to basal/mesenchymal, implicating the JAK2/STAT3 signaling axis as a key mediator of chemotherapy resistance[9]. These findings underscore the power of scRNA-seq in uncovering distinct cell states and gene expression profiles driving drug response heterogeneity, enabling the discovery of novel biomarkers and informing personalized therapeutic strategies.

While scRNA-seq offers invaluable insights at single-cell resolution, its high dimensionality, noise, variability, and data sparsity pose significant challenges for interpretation and the extraction of meaningful biological signals. Additionally, differences in sample preparation, library construction, and sequencing platforms across laboratories introduce substantial batch effects and technical variability. To address these issues, recent advancements have introduced large-scale foundation models tailored for single-cell data analysis[14-16]. Prominent examples include scFoundation[15], scBERT[14], scGPT[16], and Geneformer[17]. These models were pre-trained on large-scale scRNA-seq datasets without being specifically designed for downstream tasks[18]. Through transfer learning and fine-tuning, single-cell foundation models have demonstrated strong performance across diverse applications, including cell type annotation and batch effect correction[18]. Notably, certain models offer specialized capabilities—scFoundation for drug response prediction[15], scGPT for multi-omics integration (e.g., gene expression and chromatin accessibility)[16], and Geneformer for predicting gene dosage sensitivity and chromatin dynamics[17]. On the other hand, general domain large language models have also been adapted to analyze sequencing data[19, 20]. Despite these advancements, no comprehensive benchmarking studies have rigorously evaluated these foundation models for drug response prediction across diverse single-cell datasets. This gap motivates our study, which aims to develop a unified platform and systematically evaluate existing large models for drug response prediction across a wide range of cancer types, tissues, and therapeutic regimens.

To achieve this, we developed scDrugMap, a unified framework for drug response prediction, and conducted the first comprehensive benchmarking of ten foundation models, comprising eight domain-specific models for single-cell data and two general-purpose natural language models. The curated dataset of scDrugMap includes a primary collection of 326,751 cells derived from 36 single-cell datasets across 23 studies, along with an external validation dataset, which includes 18,856 cells from 17 single-cell datasets across six drug-related studies. By covering a total of 345,607 single cells, we performed two evaluation strategies: pooled-data evaluation and cross-data evaluation. Our results indicate that different foundation models exhibit varying capabilities in drug response prediction, with pooled-data evaluation demonstrating better performance than cross-data evaluation. To facilitate broader application, scDrugMap is developed with a command-line tool and an interactive web server (https://scdrugmap.com/) for seamless drug response prediction using a variety of foundation models. Collectively, our study makes several key contributions: (1) offering model selection guidance based on dataset-specific characteristics and evaluation scenarios; (2) providing an integrated, practical, and interactive tool for drug response prediction, biomarker discovery, and resistance mechanism analysis; (3) establishing a benchmark framework to inform and accelerate future research in drug discovery.

## RESULTS

### Overview of scDrugMap

As an integrated framework for drug response prediction using single-cell data (**Fig. 1a**), scDrugMap consists of three core components: a comprehensive computational pipeline, an interactive web server, and a large-scale curated drug-related dataset—all designed to enable streamlined user access and efficient model evaluation. At the core of scDrugMap are foundation models (FMs), including eight single-cell-specific foundation models (tGPT[21], scBERT[14], Geneformer[17], cellLM[22], scFoundation[16], scGPT[16], cellPLM[23], and UCE[24]) and two general natural language models (LLaMa3-8B[25] and GPT4o-mini[26]). To adapt these models for drug response prediction, scDrugMap supports both zero-shot inference and fine-tuning using Low-Rank Adaptation (LoRA). Users can interact with scDrugMap through either a command-line interface or an intuitive web-based platform, allowing flexible model experimentation and analysis. For benchmarking and validation, we manually curated two distinct single-cell datasets: a primary collection and a validation collection (*Data Collection and Preprocessing* in **Methods**). By integrating data curation, model adaptation, and predictive analysis, scDrugMap serves as a comprehensive and accessible resource for advancing single-cell-based drug response research.

As shown in **Fig. 1b**, the curated datasets in scDrugMap span 14 cancer types, 3 therapy types, 5 tissue types, and 21 treatment regimens. The primary data collection includes 36 scRNA-seq datasets manually curated from 23 published studies (**Fig. 1c**), covering 11 major cancer types such as lung cancer, multiple myeloma, and melanoma, along with three therapy categories: targeted therapy, chemotherapy, and immunotherapy. The dataset encompasses four major tissue types: cell line, bone marrow aspirates, tumor tissue, and peripheral blood mononuclear cells (PBMCs) (see **Supplementary Table 1**). Following quality control, the primary dataset comprised 326,751 single tumor cells annotated with drug response information. Among these, cell lines, targeted therapy, and lung cance represented the largest cell counts within their respective categories. Importantly, most subgroups (e.g., those for targeted therapy, bone marrow aspirates, and multiple myeloma) exhibited balanced distributions between drug-sensitive and drug-resistant cells (**Supplementary Table 2**).

For the validation data collection, we manually collected 17 scRNA-seq datasets from six other drug-related studies on solid tumors available in the GEO database[7, 8, 10-12, 27] (**Fig. 1d**). This validation dataset includes five cancer types: ovarian cancer, NSCLC, pancreatic cancer, colon cancer, and basal cell cancer, along with three therapy types: targeted therapy, chemotherapy, and immunotherapy. The represented tissue types include cell lines, tumor tissue, and organoids (**Supplementary Table 1**). After quality control, the validation set comprised 18,856 single-cell transcriptomes. Similar to the primary dataset, the largest proportions were again observed in cell line (tissue), targeted therapy (treatment), and NSCLC (cancer type) categories. Most validation subgroups also maintained a balanced representation of drug-sensitive and drug-resistant cells (**Supplementary Table 2**). Using these curated data, we implemented two distinct evaluations within scDrugMap: 1) pooled-data evaluation, where models were trained and tested on aggregated data from multiple studies; 2) cross-data evaluation, where models were tested independently on datasets from individual studies.

**Pooled-Data evaluation in primary data collection**

First, we evaluated different FMs in scDrugMap using frozen layer training for predicting drug responses in the primary data collection. The results across various tissue types, cancer types, drug classes, and regimen are shown in **Fig. 2a** and **Supplementary Fig. 1a**. Specifically, for the cell line data, which contained the highest number of cells, scFoundation demonstrated the best performance (mean F1 score: 0.971) while scBERT performed the worst (mean F1 score: 0.630). In prostate and pancreatic cancer, LLaMa3 exhibited comparable performance with scFoundation (mean F1 scores: scFoundation-0.990, LLaMa3-0.913 for prostate; scFoundation–1.000, LLaMa3–0.963 for pancreatic cancer). Additionally, LLaMa3 was the best method for predicting drug response in the CAR-T regimen (mean F1 score: 0.875), and scGPT was the best for the carboplatin regimen (mean F1 score: 0.882). Other metrics-including AUROC, accuracy, precision, and recall (**Supplementary Fig. 1a**), further demonstrate that scFoundation achieves superior drug response predictions in most of the datasets among the primary data collection. In contrast, some models demonstrated poor performance in specific contexts. For instance, CellLM and scBERT underperformed in on peripheral blood mononuclear cells data(mean F1 score: 0.461 and 0.483, respectively) while Geneformer presented poor predictions on immunotherapy datasets (mean F1 score: 0.442).

We next evaluated different models using fine-tuning layer training for predicting drug responses (**Fig. 2b**; **Supplementary Fig. 1b**). The results from fine-tuning were overall consistent with those from embedding-based approaches, with scFoundation remaining the best and scBERT remaining the worst across most of the datasets. An exception was observed with melanoma cancer data, where scGPT outperformed scFoundation (mean F1 scores: 0.993 vs 0.978). In liver cancer, cellPLM performed comparably to scFoundation (mean F1 scores, cellPLM: 0.992; scFoundation: 0.997), both outperforming other models. Across drug regimens, scGPT performed similarly to scFoundation for ceritinib (mean F1 scores, scGPT: 0.985; scFoundation: 0.997) and outperformed scFoundation on vemurafenib (mean F1 scores, scGPT: 1.000; scFoundation: 0.990). Similarly, cellPLM demonstrated comparable performance to scFoundation for ceritinib (mean F1 scores, cellPLM: 0.990; scFoundation: 0.997), sorafenib (mean F1 scores, cellPLM: 0.996; scFoundation: 0.999), and vemurafenib (mean F1 scores, cellPLM: 0.977; scFoundation: 0.990). Other metrics, including AUROC, accuracy, precision, and recall (**Supplementary Fig. 1b**), showed that scFoundation achieves accurate response predictions in diverse datasets.

Unlike the results under the layer-freezing strategy, all models in the fine-tuning setting outperformed the baseline models across various tissue types. Notably, scFoundation achieved the highest mean F1 scores across all tissue categories: 0.940 for peripheral blood mononuclear cells, 0.990 for tumor tissue, 0.962 for bone marrow aspirates, and 0.947 for cell lines (**Fig. 2b**). Additionally, UCE showed strong performance specifically in tumor tissue analysis, achieving a mean F1 score of 0.869. When evaluating performance across different cancer types, scFoundation remained the top-performing model in all cases except for melanoma, where scGPT outperformed it with a mean F1 score of 0.992, compared to 0.978 for scFoundation. In terms of drug type categories, scFoundation consistently demonstrated the best performance, with mean F1 scores of 0.953 for targeted therapy, 0.996 for chemotherapy, and 0.832 for immunotherapy.

Taken together, fine-tuning the FMs generally outperformed layer-freezing based prediction. Among the evaluated models in scDrugMap, scFoundation, when fine-tuned, consistently demonstrated superior performance in drug response prediction across diverse datasets, including various tissues, drugs, cancer types, and regimens. scGPT, also fine-tuned, excelled in predicting responses for melanoma cancer and the vemurafenib regimen. In contrast, cellLM particularly underperformed in immunotherapy and neuroblastoma, under both training strategies.

**Cross-data evaluation in primary data collection**

To assess the generalizability of FMs in scDrugMap, we performed cross-data evaluation, in which models were trained on certain studies and tested on independent, unseen datasets. This evaluation simulates de novo drug response prediction scenarios in real-world settings.

We first examined model performance using layer-freezing training. In general, most models performed less effectively across various tissue types, drug classes, cancer types, and regimens, compared to the pooled-data evaluation results (**Fig. 3a**; **Supplementary Fig. 2**). Nevertheless, a few models showed promising results. Specifically, regarding tissue types, scGPT achieved a mean F1 score of 0.858 in tumor tissue, UCE scored 0.723 in tumor tissue, and cellPLM reached 0.676 in bone marrow aspirate. For drug type, chemotherapy showed the best performance, with UCE and cellLM achieving mean F1 scores of 0.577 and 0.617, respectively. Regarding regimens, the top three performances were observed for UCE and scBERT with ibrutinib (mean F1 scores, UCE: 0.645; scBERT: 0.570), and scGPT with paclitaxel (mean F1 score: 0.566). Across cancer types, LLaMa3 generally showed the lowest variance in F1 scores. scFoundation achieved the highest performance in multiple myeloma, with a mean F1 score of 0.734. cellLM performed best specifically in acute lymphoblastic leukemia (mean F1 score: 0.635) and prostate cancer (mean F1 score: 0.636).

Next, we evaluated model performance under the fine-tuning strategy, where UCE achieved optimal results across various categories (**Fig. 3b**; **Supplementary Fig. 3**). For example, it was the best model across tissue types (highest mean F1 score of 0.681 in bone marrow aspirate), drug types (highest mean F1 score of 0.549 in targeted therapy), and regimens (highest mean F1 score of 0.677 in paclitaxel). In contrast, scFoundation, which had performed best under pooled-data evaluation, showed reduced generalizability in the cross-data setting. It was notably the least effective model for the ibrutinib regimen. Interestingly, cellPLM ranked second across drug types (mean F1 score of 0.681 in immunotherapy). Across cancer types, UCE showed top performance in acute lymphoblastic leukemia (ALL, mean F1 score of 0.752), while scFoundation maintained strong results in multiple myeloma (mean F1 score of 0.743). Additional strong performers included tGPT and cellLM for ALL (mean F1 scores: tGPT-0.728; cellLM-0.718).

To further interpret model behavior, we visualized model-specific embeddings using Uniform Manifold Approximation and Projection (UMAP). These visualizations revealed that scFoundation, UCE, and scGPT effectively distinguished drug-resistant and drug-sensitive cells, while other models showed limited separation (**Fig. 4**). Collectively, under the cross-data evaluation scenario, most models yielded comparable performance by layer-freezing and fine-tuning, although both showed generally poorer performance than the pooled-data evaluation scenario. When using layer-freezing, scGPT outperformed others in tumor tissue data, while under fine-tuning, UCE consistently delivered the best performance across various tissue types, drug categories, cancer types, and regimens.

**Few-shot learning for drug response prediction**

While single-cell FMs such as scFoundation[15], UCE[24], and scGPT[16] demonstrated strong performance in predicting drug responses from single-cell transcriptomic data, we also included a general-purpose foundation model, GPT4o-mini, in the scDrugMap and investigated its applicability in this context. Unlike scFMs specifically pre-trained on single-cell or biomedical data, GPT4o-mini was developed for general natural language tasks. We aimed to assess its zero-shot reasoning capabilities when prompted with single-cell gene expression features for binary drug sensitivity classification (**Fig. 5a**).

Overall, GPT4o-mini exhibited limited predictive performance, performing near or below baseline levels in most evaluation settings. The model's highest performance was observed in liver cancer, achieving a mean F1 score of 0.690, which suggests some potential context-specific utility. However, across most other categories, its predictive accuracy was substantially lower. In terms of tissue types, GPT4o-mini achieved mean F1 scores of 0.583 in peripheral blood mononuclear cells, 0.507 in tumor tissue, 0.479 in cell lines, and 0.459 in bone marrow aspirates, highlighting a clear performance gap compared to specialized scFMs. Across drug classes, the model performed best on targeted therapy (mean F1: 0.538), followed by chemotherapy (0.517) and immunotherapy, where performance was notably poor (0.391). When stratified by cancer type, GPT4o-mini's performance varied widely. Besides liver cancer, the model showed modest performance for chronic lymphocytic leukemia (0.622) and lung cancer (0.594), but struggled on others such as pancreatic cancer (0.190), melanoma (0.320), and breast cancer (0.387) (**Fig. 5b**). These results highlight that, without domain-specific pretraining or adaptation, general-purpose language models lack the precision and robustness required for reliable drug response prediction in single-cell settings.

**Model evaluation in the validation data collection**

To further assess model generalizability, we evaluated the performance of foundation models in scDrugMap using a separate validation dataset collection (**Fig. 6; Supplementary Fig. 4**). For this analysis, we applied the pooled-data evaluation strategy, where all validation datasets were aggregated for model training and testing. Among all models, scFoundation demonstrated the best performance across all categories, including tissue types (mean F1 scores, cell line: 0.901; tumor tissue: 0.965; organoid: 0.951), drug types (mean F1 scores, targeted therapy: 0.946; chemotherapy: 0.900; immunotherapy: 0.918), and cancer types (mean F1 scores, NSCLC: 0.997; ovarian cancer: 0.920; basal cell cancer: 0.918; pancreatic: 0.951; colon cancer: 0.847) (**Fig. 6a**). Additionally, This strong performance was further supported by AUROC metrics, where scFoundation again led across tissue types, drug types, and cancer types (**Fig. 6b**). Other metrics including accuracy, precision, and recall also showed that scFoundation provides accurate predictions in this validation data collection.

In addition to scFoundation, both scGPT and tGPT demonstrated competitive performance in selected categories. For example, across tissue types, they performed well in tumor tissue (mean F1 scores, scGPT: 0.942; tGPT: 0.944) and organoid (mean F1 scores, scGPT: 0.876; tGPT: 0.909), but not in cell line (mean F1 scores, scGPT: 0.560; tGPT: 0.740) (**Fig. 6a**). Across drug types, the two models performed similarly, with mean F1 scores ranging from 0.714 to 0.890. In terms of cancer types, cellPLM, scGPT, and tGPT displayed similar performance, performing well across most cancers (mean F1 scores ranged from 0.755 to 0.986) except for ovarian cancer, where scGPT yielded a lower score (0.578) compared to cellPLM (0.700) and tGPT (0.753).

**Computational scalability**

In addition to prediction accuracy, scDrugMap also assesses the computational scalability of each foundation model by comparing their architectures, parameter counts, output dimensions, and runtime efficiency (**Fig. 7**). These factors are critical for selecting models that balance performance with practical deployment needs in large-scale single-cell analysis.

For example, scFoundation, using an encoder-decoder architecture, stood out with its large parameter count (121.2 million), high output dimension of 3,072, and efficient training and inference speeds (23.26 it/s and 69.98 it/s, respectively), making it one of the most efficient and powerful models in our benchmark. In comparison, scGPT, which also adopts a dual-architecture approach, had fewer parameters (52.5 million) and a smaller output dimension (512), with lower efficiency (1.43 it/s for training; 1.44 it/s for inference) than scFoundation. Another encoder-decoder model, UCE, offers competitive inference and training speeds, though it has a smaller output dimension of 1280. On the other hand, models like tGPT (decoder-only), scBERT (encoder-only), and Geneformer (encoder-only) used a single architecture and are relatively less efficient, with lower output dimensions (1024 for tGPT, 200 for scBERT, 256 for Geneformer) and slower training and inference speeds. Taken together, these results highlight important trade-offs between architectural complexity, model size, and computational performance. By incorporating these metrics, scDrugMap enables users to make informed decisions about model selection based not only on accuracy but also on runtime efficiency and resource constraints, tailoring deployment to the specific demands of different single-cell datasets and drug response tasks.

## DISCUSSION

In this study, we introduce scDrugMap, a comprehensive framework for benchmarking large-scale foundation models in single-cell drug response prediction. We evaluated eight single-cell foundation models and two general-purpose language models using a primary dataset of 326,751 cells and a validation set of 18,856 cells, spanning five tissue types, three drug classes, fourteen cancer types, and twenty-one treatment regimens. Models were assessed under two strategies—pooled-data and cross-data evaluation—and three training approaches: layer-freezing, fine-tuning, and zero-shot learning. scFoundation consistently performed best in pooled-data evaluations, while UCE and scGPT excelled in cross-data scenarios. General-purpose models like LLaMa3-8B and GPT4o-mini showed limited performance compared to single-cell-specific models.

Prior to the rise of foundation models, statistical and deep learning (DL) frameworks[28] were widely used for single-cell drug response prediction[29]. For example, DREEP is a tool that relies on well-established publicly available pharmacogenomic profiles and functional enrichment analysis to predict an individual cell's vulnerability to hundreds of drugs[30]. Chen et al. introduces scDEAL, a deep transfer learning model that integrates scRNA-seq data with drug-response annotated bulk RNA-seq to transfer the bulk-level knowledge to the single-cell level[31]. Another study, DeepDR, provides a DL library designed for drug response prediction[32]. It supports 135 types of DL models by incorporating a wide range of drug and cell features, along with multiple encoders and fusion modules[32]. These tools are often specialized for specific biological tasks and were trained on domain-specific data, such as the Genomics of Drug Sensitivity in Cancer (GDSC)[33], Cancer Therapeutics Response Portal (CTRP) (http://www.broadinstitute.org/ctrp/), Pharmacogenomics of Responses in Cancer Cell Lines (PRISM) (https://depmap.org), and the Cancer Cell Line Encyclopedia (CCLE)[34]. Although these frameworks can offer great performance for targeted tasks, they require extensive domain-specific training to improve stability and robustness and may face challenges with scalability and generalization.

In contrast, large-scale, pre-trained foundation models are built on broader knowledge and can transfer learnings across different downstream tasks. Single-cell foundation models, such as scFoundation, scGPT, and scBERT are trained on millions of scRNA-seq data with a variety of cell types from different sources. These models demonstrate the ability to generalize across multiple datasets and tasks without requiring extensive retraining on, for example, specific drug-response data. Foundation models thus represent a promising shift towards more flexible, scalable, and data-efficient approaches in single-cell drug response prediction. However, the advantage of foundation models in this domain is not without limitations. Despite their high computational efficiency and generalizability, our evaluation in scDrugMap demonstrated that fine-tuning consistently yielded better performance than layer-freezing based prediction, indicating that pretrained models still benefit from exposure to task-specific data, particularly in capturing the complex relationships between cellular states and drug responses.

Additionally, our study highlighted that models in pooled-data evaluation outperformed that of cross-data evaluation. That indicates that single-cell foundation models benefit from exposure to a broader range of data within a given category. When datasets from multiple studies within the same category (e.g., tumor tissue) are pooled together, the broader diversity within the pooled datasets may allow the model to learn from various sources for more accurate prediction. In contrast, when training is performed on each dataset separately, the model may struggle to generalize, as it is less able to apply its pre-trained knowledge to a narrower and more specific context. Thus, large-scale pooled data helps ensure the foundation model can apply its broad pre-trained knowledge more effectively, resulting in more robust drug response predictions.

Looking ahead, several future directions for single cell foundation models in drug response prediction can be envisioned. First, there remains a need for the incorporation of domain-specific knowledge into model training. Second, integrating multi-modal data sources, such as transcriptomic, proteomic, and genomic data, to enhance the models' ability to predict drug responses more accurately across diverse patient populations. Third, further exploration of hybrid models combining the strengths of single-cell foundation models and natural language models may lead to more powerful and generalized prediction tools that can effectively handle the complexities of single-cell data while benefiting from broader biological knowledge. Finally, incorporation of explainable AI techniques to enhance model interpretability and biological relevance of predictions made by foundation models, providing clearer insights into the biological mechanisms behind drug responses.

## METHODS

### Overview of scDrugMap Framework for Drug Response Prediction

In the scDrugMap framework, we identified recently developed single-cell-related foundation models (scFMs) for drug response prediction. In total, we selected and evaluated ten FMs, comprising eight scFMs and two general-purpose natural language models. The scFMs evaluated in this study include transcriptome-GPT[35] (tGPT), single-cell Bidirectional Encoder Representations from Transformers[14] (scBERT), Geneformer[17], Single-Cell Language Model[22] (CellLM), single-cell Foundation Model[15] (scFoundation), single-cell Generative Pre-Trained Transformer[16] (scGPT), Cell Pre-training Language Model[23] (CellPLM), and Universal Cell Embeddings[24] (UCE). The two natural language models are Llama3-8B[25] and GPT4o-mini[26].

### Model Evaluation Settings

In scDrugMap, we evaluated the FMs in two scenarios: the pooled-data scenario and the cross-data scenario. In the pooled-data scenario, datasets from multiple studies within the same category (e.g., tissue type, drug, regimen, or cancer type) were combined into a single dataset. This pooled dataset was then randomly split into training and test sets, allowing the model to learn and be evaluated on data drawn from the same mixed source. In contrast, the cross-data scenario tested the model's ability to generalize across studies. The model was trained on data from one or more studies and evaluated on data from a completely separate study. In this setting, training and testing were performed using data from different studies, while still remaining within the same category (e.g., tissue type, drug, regimen, or cancer type). This setting can be considered a zero-shot capability evaluation of the model, which better reflects real-world conditions where models are often applied to new, unseen datasets.

**Model Training Strategies**

For both evaluation scenarios, we applied two training strategies: layer-freezing and LoRA[36] (Low-Rank Adaptation). These strategies were used to train the foundation models (FMs) on the training sets and evaluate their performance on corresponding test sets. For LLaMa3-8B, only the layer-freezing strategy was applied. For GPT4o-mini, we employed a prompt engineering approach similar to that used in GPTCelltype[37], without additional fine-tuning.

In the layer-freezing strategy, pretrained scFMs treat each gene (by ENSEMBL ID or gene symbol) as an individual token within a fixed vocabulary (dictionary). To extract feature embeddings for downstream classification, we averaged all token embeddings generated by the scFM. These aggregated embeddings were then passed to a multi-layer perceptron (MLP) for classification. The MLP structure used for classification was defined as follows:

$$\boldsymbol{h}_1 = ReLU(\boldsymbol{W}_1 \boldsymbol{x} + \boldsymbol{b}_1),$$

$$\boldsymbol{h}_2 = ReLU(\boldsymbol{W}_2 \boldsymbol{h}_1 + \boldsymbol{b}_2),$$

$$\widehat{\boldsymbol{y}} = Sigmoid(\boldsymbol{W}_3 \boldsymbol{h}_2 + \boldsymbol{b}_3),$$

where $\boldsymbol{x} \in \mathbb{R}^d$ is the embedding representation produced by the scFM, $d$ is the model-specific output dimension, and $\widehat{\boldsymbol{y}}$ is the predicted probability of drug sensitivity. $ReLU$ and $Sigmoid$ are the activation functions, and $\boldsymbol{W}_i$, $\boldsymbol{b}_i$ represent weights and biases of fully connected layers. We used 10-fold cross-validation, randomly partitioning the data within each biological category (e.g., tissue type, drug class, cancer type, or treatment regimen). The same data partitions were used consistently across models. Only the MLP parameters were trained, while the foundation model weights remained frozen.

In the fine-tuning strategy, we fine-tuned a subset of the model parameters using LoRA, implemented via the peft[38] package, while keeping the rest of the model frozen. LoRA enables efficient adaptation of large models by injecting low-rank matrices into trainable layers. We set the LoRA rank to 8, alpha (scaling) to 8, dropout to 0.05, and defined the task type as "SEQ_CLS" (sequence classification). The output embeddings from the partially fine-tuned model were again passed through the MLP classifier, and both the MLP parameters and LoRA-injected parameters were trained using backpropagation. Let a pretrained single cell foundation model be denoted as $scFM(D, W)$, where $W = \{\widehat{W}, W_{ft}\}$, with $\widehat{W}$ representing frozen weights and $W_{ft}$ representing the trainable layers. Given a fine-tuning dataset $D_{ft} = \{(x_i, y_i)\}_{i=1}^k$, the objective is to minimize the binary cross-entropy loss:

$$\boldsymbol{W}^* = \underset{\theta_{ft}}{arg\,min}\, E_{x,y \in D_{ft}}[L(\widehat{y}, y)],$$

$$\widehat{y} = scFM_{ft}(x; \widehat{W}, \boldsymbol{\Phi}_{ft}, W_{ft}),$$

where $\boldsymbol{\Phi}_{ft}$ are the parameters of the MLP classifier, and $L$ is the binary cross-entropy loss.

In LoRA, weight updates are expressed as a low-rank approximation:

$$W' = W_{ft} + \Delta W = W_{ft} + AB$$

, where $A \in \mathbb{R}^{d \times r}$ and $B \in \mathbb{R}^{r \times k}$ are trainable low-rank matrices, and $r$ is the rank (set to 8). Here $W_{ft}$ remains frozen and $\Delta W$ is learned during fine-tuning.

**tGPT**

tGPT takes gene expression rankings for generative pretraining on 22.3 million single-cell transcriptomic data. Gene sequence are sorted in descending order of expression, enclosed with special tokens <start> and <end>, and padded with <pad>. We followed instructions from the official tGPT GitHub repository (https://github.com/deeplearningplus/tGPT). For benchmarking, we used the pretrained model 'transcriptome-gpt-1024-8-16-64', which is stored in https://huggingface.co/lixiangchun/transcriptome-gpt-1024-8-16-64. The input sequence length of tGPT is 64 and the output embedding dimension is 1024. For layer-freezing training, we average the output of the last layer in the sequence dimension as the input of the downstream classifier. The target module $W_{ft}$ for LoRA fine-tuning is the attention layer of Transformer block called 'c_attn'.

**scBERT**

scBERT uses gene2vec technology to embed genes into a predefined vector space, reflecting the similarity between genes and simplifying model training[14]. To effectively utilize the transcription levels of each gene, the model discretizes gene expression and converts it into a 200-dimensional vector, similar to the word frequency analysis method in NLP, which serves as the token input for the model. Moreover, scBERT adopts the Performer architecture to enhance limitation of input length and model efficiency. Though scBERT can handle very long input sequences, we chose to use the first 8,000 tokens to maintain uniformity in batch operations. For drug response prediction, we used the pre-trained model provided at https://github.com/TencentAILabHealthcare/scBERT. For the choice of layer-freezing embedding, we keep it consistent with the original operation of scBERT and average the output of the last layer in the sequence dimension. LoRA fine-tuning $W_{ft}$ includes the key ('to_k'), value ('to_v') and query ('to_q') matrices of the model.

**Geneformer**

Geneformer consists of six standard Transformer encoder blocks. The model features an input size of 2,048, an embedding dimension of 256, four attention heads per layer, and a feedforward layer of size 512. Geneformer is pre-trained on 30 million cells and released two versions of the pre-trained model. For this drug response prediction task, we used the pre-trained model named as 'geneformer-6L-30M_CellClassifier_cardiomyopathies_220224', which is available and can be obtained from https://huggingface.co/ctheodoris/Geneformer/tree/main. For the input, we follow the same approach as Geneformer by first sorting gene expression values in descending order and then retaining the top 2,048 genes. For drug response prediction using fixed-layer embeddings, we utilized the representations extracted from the 'hidden_states' output of the final Transformer encoder layer. For LoRA fine-tuning, the target module $W_{ft}$ included the key ('key'), value ('value"), and query ('query') matrices.

**CellLM**

Traditional methods rely on the BERT architecture, which leads to anisotropy in the embedding space and inefficient semantic representation. CellLM addresses the computational limitations caused by large batches of data through a new divide-and-conquer contrastive learning method, thereby improving the quality of representation. Trained on 2 million scRNA-seq data, CellLM is the first large language model that learns from both normal cells and cancer cells. To ensure batch consistency, we also limit the max input length to 8000. We choose to take the average of the hidden states output by the encoder in the sequence dimension as the layer-freezing embedding. CellLM utilizes a Performer structure similar to scBERT, so the components involved in $W_{ft}$ are consistent with scBERT, specifically the key ('to_k'), value ('to_v') and query ('to_q') matrices of the model. The pre-trained model for CellLM can be accessed at https://github.com/PharMolix/OpenBioMed/tree/main.

**scFoundation**

scFoundation has more than 100 million parameters and is trained on over 50 million human single-cell transcriptome data, covering the complex molecular features of all known cell types[15]. In addition to gene expression values, the input of scFoundation includes two additional indicators: 'S' (source) and 'T' (target), which represent the input value and the total gene expression, respectively. scFoundation adopts an encoder-decoder design, using only non-zero genes during the encoding stage, while both expressed and non-expressed genes are considered during the decoding stage. We used the pretrained model available at https://github.com/biomap-research/scFoundation. For fixed-layer drug response prediction, the cell embedding was constructed by concatenating four componennts, as described in the original paper: the representation of 'S' and 'T', along with the maximum and average embeddings of all genes. LoRA fine-tuning was applied to the model's output layer.

**scGPT**

scGPT is pretrained on over 33 million single-cell sequencing data and is further optimized through transfer learning[16]. For its input, scGPT not only uses information about genes and expression values, but also introduces the special Condition Tokens. The Condition Tokens encompass diverse meta information associated with 502 individual genes, such as modality, batch, and perturbation conditions. Additionally, scGPT can handle unknown genes to address the Out of Vocabulary problem. The model can be accessed at https://github.com/bowang-lab/scGPT. The cell embedding used for layer-freezing based drug response prediction is the <CLS> token (the token defined before pre-training, its representation will be trained during the training phase, and is usually added as the first token in the sequence), which can represent the entire sentence. For LoRA fine-tuning, it targeted the output layer ('out_proj').

**CellPLM**

CellPLM addresses the differences between single-cell data and natural language data that are often overlooked by existing models. Unlike traditional methods that treat genes as tokens and cells as sentences, CellPLM treats cells as tokens and tissues as sentences, using spatial transcriptome data during pre-training to better learn the relationships between cells. They consider the embedding at l-th layer of Transformer as a set of N tokens, where N is the total number of cells in a tissue sample. Additionally, CellPLM introduces a Gaussian mixture prior distribution to mitigate issues of insufficient data and noise. The pre-trained model we used is '20230926_85M' from https://github.com/OmicsML/CellPLM. The cell embedding used for layer-freezing drug response prediction is the output of the last layer in CellPLM. LoRA fine-tuning includes the query, key, and value ('query_projection', 'key_projection', and 'value_projection') matrices in the CellPLM model.

**UCE**

UCE provides a unified biological latent space through fully self-supervised training with single-cell transcriptome data[24]. The UCE is capable of capturing significant biological changes and can map any new cell to this embedding space without additional data annotation, model training, or fine-tuning. UCE employs an encoder-decoder structure and performs specific processing on the input, such as adding special tokens (e.g., the CLS token is designed to capture cell-level embeddings when training the model). We used the UCE model provided at https://github.com/snap-stanford/UCE. For layer-freezing drug response prediction, the embeddings was taken from the <CLS> token (the first token in the sequence), as provided by UCE. For LoRA fine-tuning, the output layer out_proj was designated as the target module.

**LLaMa3-8B**

In addition to the above-mentioned scFMs, we also tested the performance of general FMs in this task, specifically using the open-source LLM released by Meta, Llama-3-8B. Llama-3-8B has nearly 8 billion parameters and is designed to provide efficient natural language understanding and generation capabilities. It is an upgraded version of the Llama series, capable of handling more complex language tasks while optimizing inference speed and resource efficiency. For input processing, to align with the methods used by most scFMs, we first arranged the genes in the sequence from high to low based on expression levels. We then limited the input length to 1,024 genes, treating these gene symbols as natural language (separated by space) and used Llama-3-8B's tokenizer for tokenization. Finally, the embedding output by Llama-3-8B was used for layer-freezing drug response prediction. The model can be obtained by applying for authorization from Meta and downloading it from https://huggingface.co/meta-llama/Meta-Llama-3-8B.

**GPT4o-mini**

Inspired by GPTCelltype[37], we also tested the latest generation of FMs, GPT4o-mini, recently released by OpenAI. It contains knowledge up to July 2024. Like GPTCelltype, we use prompt words to guide the FM's responses. The specific settings for the model input are as follows:

$$input = prompt_{pre} + prompt_{source} + \boldsymbol{source} + prompt_{gene} + \boldsymbol{content} + prompt_{post}$$

The input consists of 4 prompt modules, source, and content. The description of each prompt module is shown in **Fig. 5a**. The $\boldsymbol{source}$ is the data source of the current single-cell sequence. We hope to help obtain accurate results by prompting the FM with data source. The $\boldsymbol{content}$ is the name of 10 genes with the highest expression in the single cell sequence. To avoid the problem of hallucination caused by too long input in the large language model, we use the mini batch technique to only let the FM answer 10 sequences in each context [questions on fig.5]. Our experiments were conducted by calling the GPT4o-mini's API provided by OpenAI (https://openai.com/api).

**Data collection and preprocessing**

To systematically collect scRNA-seq data, we conducted a literature search in PubMed for studies published up until 2024, using the keywords '(drug resistance) AND ((single cell) OR (scRNA))'. Only samples from *Homo sapiens* with annotated drug response information were included. We also included available datasets collected on drug resistance mechanisms[39]. This yielded a total of 36 scRNA-seq datasets as the primary data collection, encompassing 11 cancer types, three treatment types, and four tissue types (**Supplementary Table 1**). To construct an external validation collection, we followed the same curation methodology, focusing on studies published since January 2024. This resulted in 17 scRNA-seq datasets for drug response studies on solid tumors covering five cancer types, three treatment types, and three tissue types (**Supplementary Table 1**). Both the primary and validation collections included experimental annotations of binary drug response status (sensitive/resistant) at cell-level (**Supplementary Table 2**). Specifically, cells of the nonresponsive samples collected from pre- or post-treatment conditions are labeled as resistant cells, while cells of the responsive samples are labeled as sensitive cells.

**Evaluation metrics**

We used four evaluation metrics (F1, AUROC, precision, and recall) for assessing model performance in predicting cell-level drug response. Accuracy is the proportion of correctly predicted samples out of the total number of samples and is most appropriate when the class distribution is balanced. F1 score is the harmonic mean of Precision and Recall, making it suitable for situations where class distribution is imbalanced. F1 can be expressed as: $Precision = TP/(TP + FP)$, $Recall = TP/(TP + FN)$, $F1\ Score = (2 \times Precision \times Recall)/(Precision + Recall)$, where TP (True Positive) represents the number of samples correctly predicted as positive, while TN (True Negative) denotes the number of samples correctly predicted as negative. FP (False Positive) refers to the number of samples incorrectly predicted as positive, indicating a false alarm, and FN (False Negative) is the number of samples incorrectly predicted as negative, signifying a missed detection. AUROC represents the area under the Receiver Operating Characteristic (ROC) curve, which is used to evaluate the discriminatory ability of a classification model. The ROC curve plots the True Positive Rate (TPR) against the False Positive Rate (FPR), where $TPR = TP/(TP + FN)$, $FPR = FP/(FP + TN)$, $AUROC = \int_0^1 TPR(FPR)dFPR$. TPR (True Positive Rate) is also known as Sensitivity or Recall. FPR is False Positive Rate. The closer the AUROC value is to 1, the better the model's ability in predicting accurate drug response.

**Web server implementation**

We developed scDrugMap as a web-based system for drug response prediction and benchmarking, built on a robust three-tier architecture consisting of the client, server, and database layers. The backend database was implemented using MySQL version 3.23 with the MyISAM storage engine, selected for its simplicity and fast read performance, which suits the platform's analytical needs. The server layer was constructed using Perl and its DBI (Database Interface) module, enabling efficient interaction with the MySQL database for dynamic content retrieval and user-driven queries. The client layer provides an intuitive, user-friendly web interface designed to facilitate access to curated datasets, model results, and visualization tools. Through this architecture, scDrugMap allows users to seamlessly explore drug response predictions, compare the performance of foundation models, and interact with large-scale single-cell datasets, supporting translational research in drug discovery and precision oncology.

**FIGURE LEGEND**

**Fig. 1 Overview of the scDrugMap framework and curated datasets.** **(a)** Schematic of the scDrugMap framework, which integrates a benchmarking platform, computational pipeline, interactive web server, and curated drug response datasets. Users can select from a range of foundation models (FMs), including single-cell-specific foundation models (scFMs) and general-purpose language models, and apply different training strategies (layer-freezing or fine-tuning) to predict drug response outcomes (sensitive or resistant). **(b)** Categories used for benchmarking model performance, including tissue types, drug types, and cancer types. **(c)** Summary of the primary dataset collection, showing the number of datasets (left) and number of cells (right) across tissue types, drug types, and cancer types. **(d)** Summary of the validation dataset collection, with the number of datasets (left) and number of cells (right) across tissue types, drug types, and cancer types.

**Fig. 2 Model performance in predicting drug response in pool-data evaluation using primary single-cell data.** F1 scores across tissue, drug, and cancer types using **a)** layer-freezing and **b)** fine-tuning training method. Error bars on the bar plots / dots on the vertical line charts represent standard deviation of the mean F1 score of each method in each category.

**Fig. 3 Model performance in predicting drug response in cross-data evaluation using primary single-cell data.** F1 scores across tissue, drug, and cancer types using **a)** layer-freezing and **b)** fine-tuning training method. Radar plots represent mean F1 scores for different tissue, drug types, and regimens (the radial axis is scaled from 0-1). Violin plots represent the kernel density distribution and the box plots inside represent the median (center line), upper and lower quartiles and 1.5× the interquartile range (whiskers) for all the cancer types. In the circular bar charts, each color segment showed the mean F1 score of the corresponding category across tissue, drug type, and regimen category. In the boxplots, the middle line is the median, the lower and upper hinges correspond to the first and third quartiles, the upper whisker extends from the hinge to the largest value no further than 1.5× the inter-quartile range (IQR) from the hinge, and the lower whisker extends from the hinge to the smallest value at most 1.5× IQR of the hinge.

**Fig. 4 UMAP projection of primary single-cell data by different methods.** UMAP embeddings using layer freezing training method are shown for **a)** scFoundation **b)** scGPT and **c)** UCE, with cells colored cancer type and cell response.

**Fig. 5 Performance of GPT4o-mini with few-shots learning in pool-data evaluation using primary single-cell data. a)** Prompt display used for GPT4o-mini. For the prompt word template, we first use a technique like the thought chain to prompt the model how it should think about the output, then tell the model the data source and sequence information, and finally we repeat telling the model and give an output template to ensure the consistent of the output format. A complete input example with prompt is also showed. **b)** Radar

plots showing the mean F1 scores of GPT4o-mini in predicting drug response across different tissue, drug types, and cancer types (the radial axis is scaled from 0-1).

**Fig. 6 Model performance in predicting drug response in pool-data evaluation using validation single-cell data.** Radar plots illustrate mean **a)** F1 scores and **b)** AUROC of each model in predicting single-cell drug response using layer-freezing training method across different tissue, drug types, and cancer types (the radial axis is scaled from 0-1).

**Fig. 7 Summary of properties, computational efficiency, and scalability of each evaluated model**. Rows correspond to algorithms ordered chronologically by year and months of publication. The first three columns display model characteristics: whether it uses an encoder-decoder architecture, the type of input embeddings, and whether it is a single-cell foundation model. The next two columns present parameters and output dimensions for each model. The next set of columns show the training and inference time and speed. For each model, the color in each cell is proportional to the corresponding value (scaled between corresponding minimum and maximum values, ignoring values of the two natural language models, shown as dashes).

**Supplementary Figures and Tables**

**Supplementary Fig. 1 Model performance in predicting drug response in pool-data evaluation using the primary data collection.** AUROC, Precision, and Recall scores across tissue, drug, regimen, and cancer types using **a)** layer-freezing and **b)** fine-tuning training method. Error bars on the bar plots represent standard deviation of the mean score of each method in each category.

**Supplementary Fig. 2 Model performance in predicting drug response in cross-data evaluation using the primary data collection.** AUROC, Precision, and Recall scores across tissue, drug, regimen, and cancer types using layer-freezing. Radar plots represent mean scores for different tissue, drug types, and regimens (the radial axis is scaled from 0-1). Violin plots represent the kernel density distribution and the box plots inside represent the median (center line), upper and lower quartiles and 1.5× the interquartile range (whiskers) for all the cancer types.

**Supplementary Fig. 3 Model performance in predicting drug response in cross-data evaluation using the primary data collection.** AUROC, Precision, and Recall scores across tissue, drug, regimen, and cancer types using LoRA fine-tuning. Radar plots represent mean scores for different tissue, drug types, and regimens (the radial axis is scaled from 0-1). Boxplots represent scores where the middle line is the median, the lower and upper hinges correspond to the first and third quartiles, the upper whisker extends from the hinge to the largest value no further than 1.5× the inter-quartile range (IQR) from the hinge, and the lower whisker extends from the hinge to the smallest value at most 1.5× IQR of the hinge.

**Supplementary Fig. 4 Model performance in predicting drug response using the validation data collection.** AUROC, Precision, and Recall scores across tissue, drug, regimen, and cancer types using layer-freezing are showed by radar plots (the radial axis is scaled from 0-1).

**DATA AVAILABILITY**

All data and results can be downloaded from the scDrugMap web server (https://scdrugmap.com). Details of metadata can be accessed from **Supplementary Table 1** and **Supplementary Table 2**.

## CODE AVAILABILITY

scDrugMap is provided as a Python package available at https://github.com/QSong-github/scDrugMap with detailed tutorials. The web server scDrugMap is available at https://scdrugmap.com and enables users to predict scRNA-seq drug response using different models.

## ACKNOWLEDGEMENTS

Q.S. is supported by the National Institute of General Medical Sciences of the National Institutes of Health (R35GM151089). G.W. is supported by the National Institute of General Medical Sciences of the National Institutes of Health (1R35GM150460). This work partially used Jetstream2[40] through allocation CIS230237 from the Advanced Cyberinfrastructure Coordination Ecosystem: Services & Support (ACCESS)[41] program, which is supported by National Science Foundation grants #2138259, #2138286, #2138307, #2137603, and #2138296.

## ETHICS DECLARATIONS

Competing interests

The authors declare no competing interests.

Fig. 1

a.

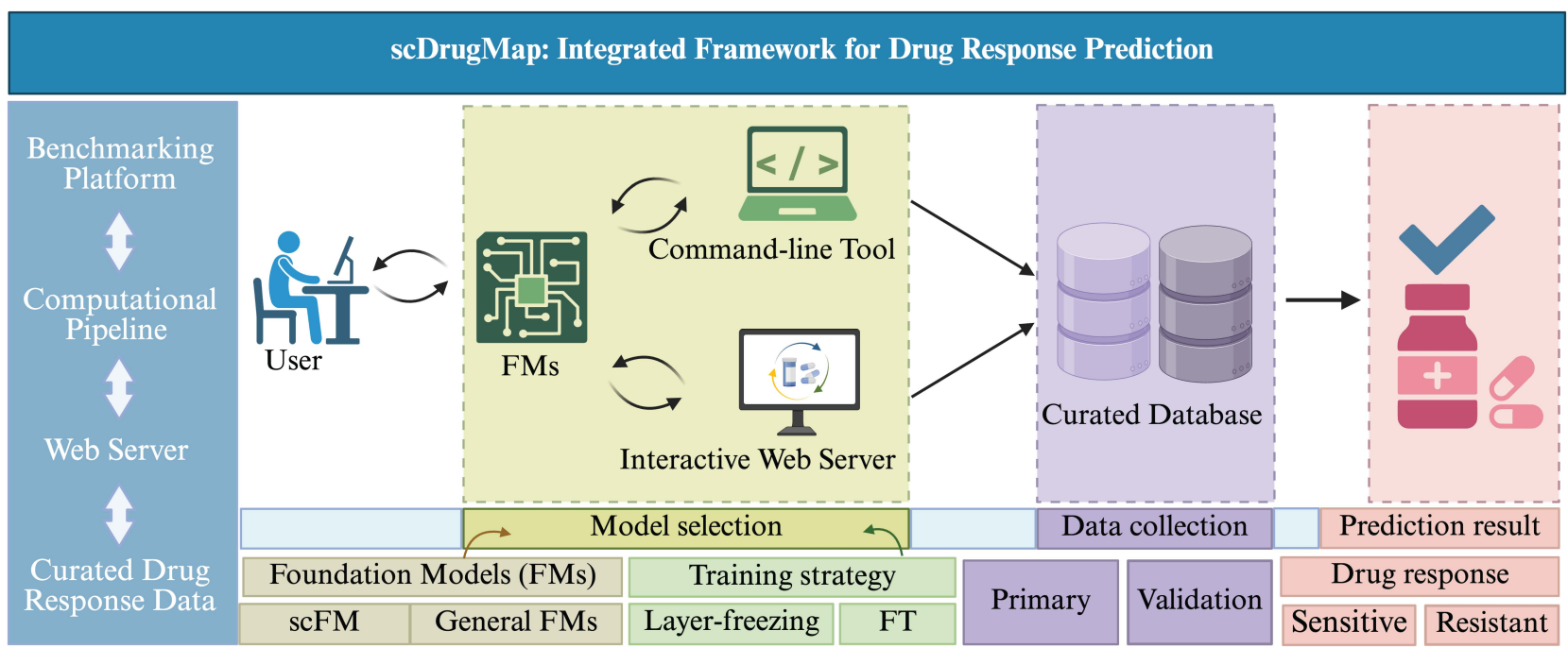

b.

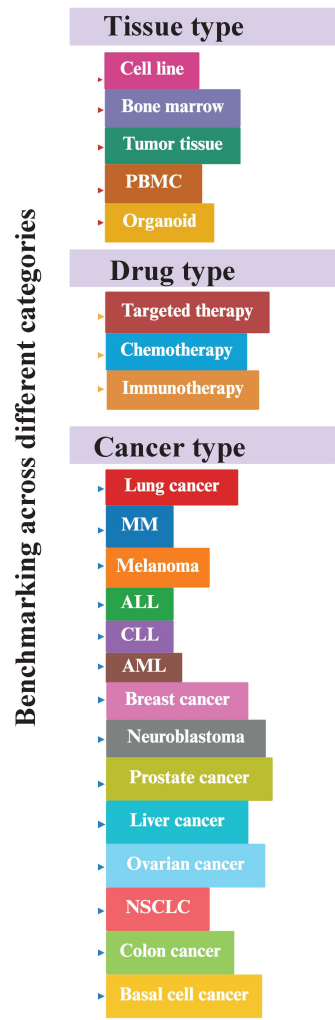

c.

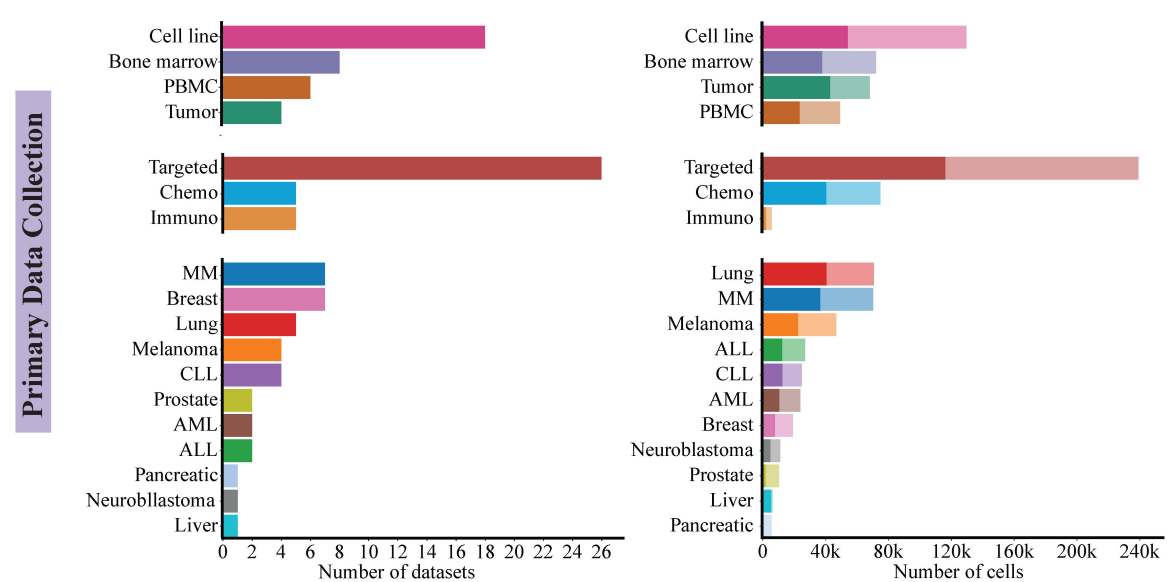

d.

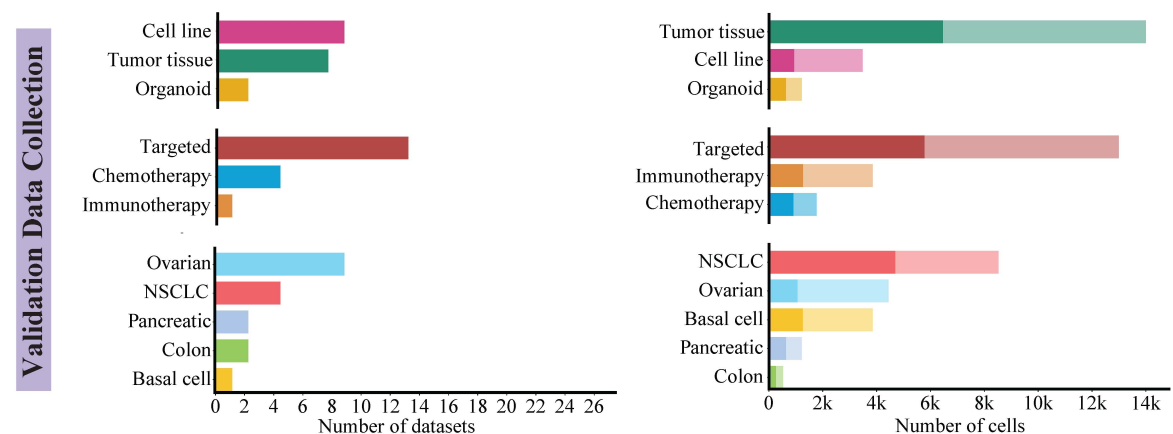

**Fig. 2**

## a. Results of Layer-freezing in pooled-data scenario

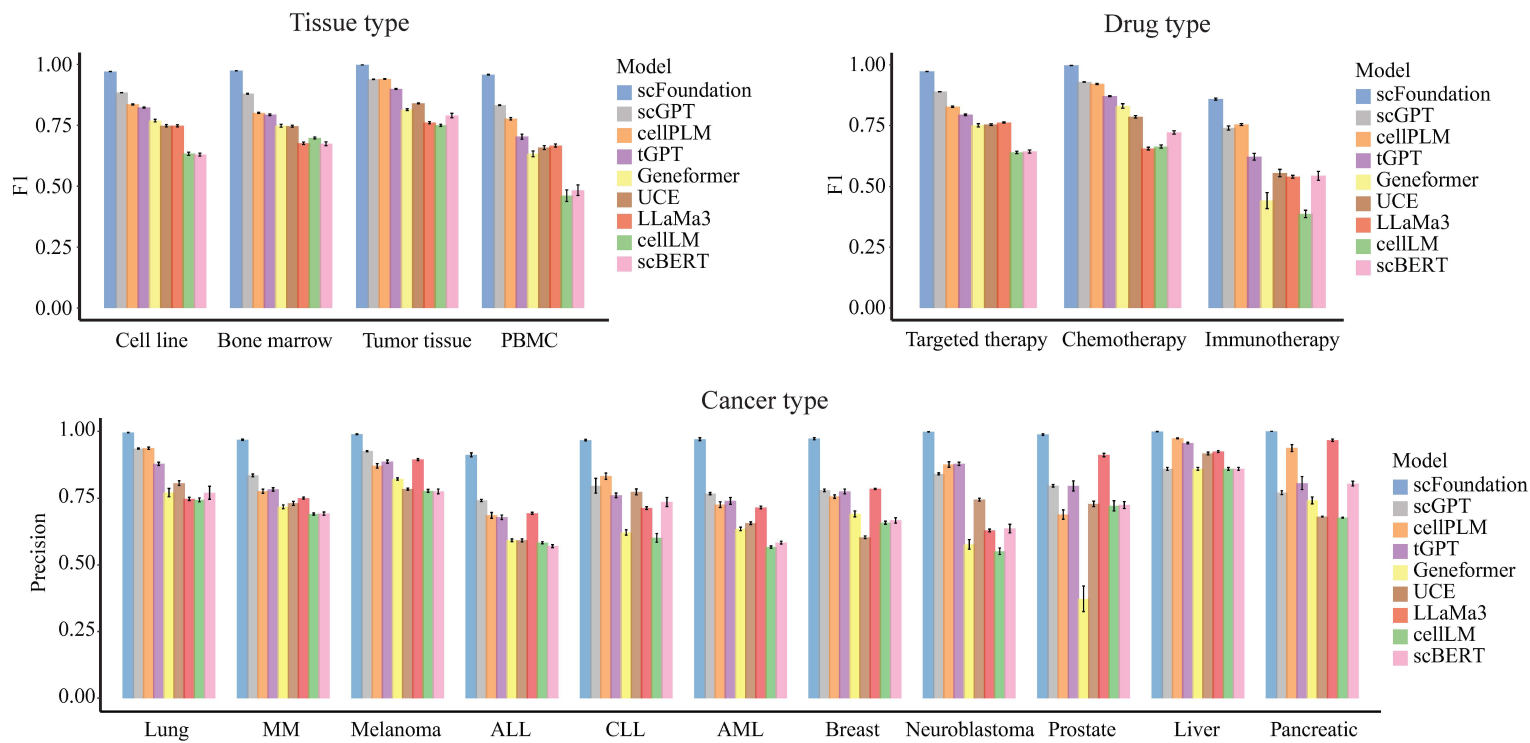

## b. Results of Fune-tuning in pooled-data scenario

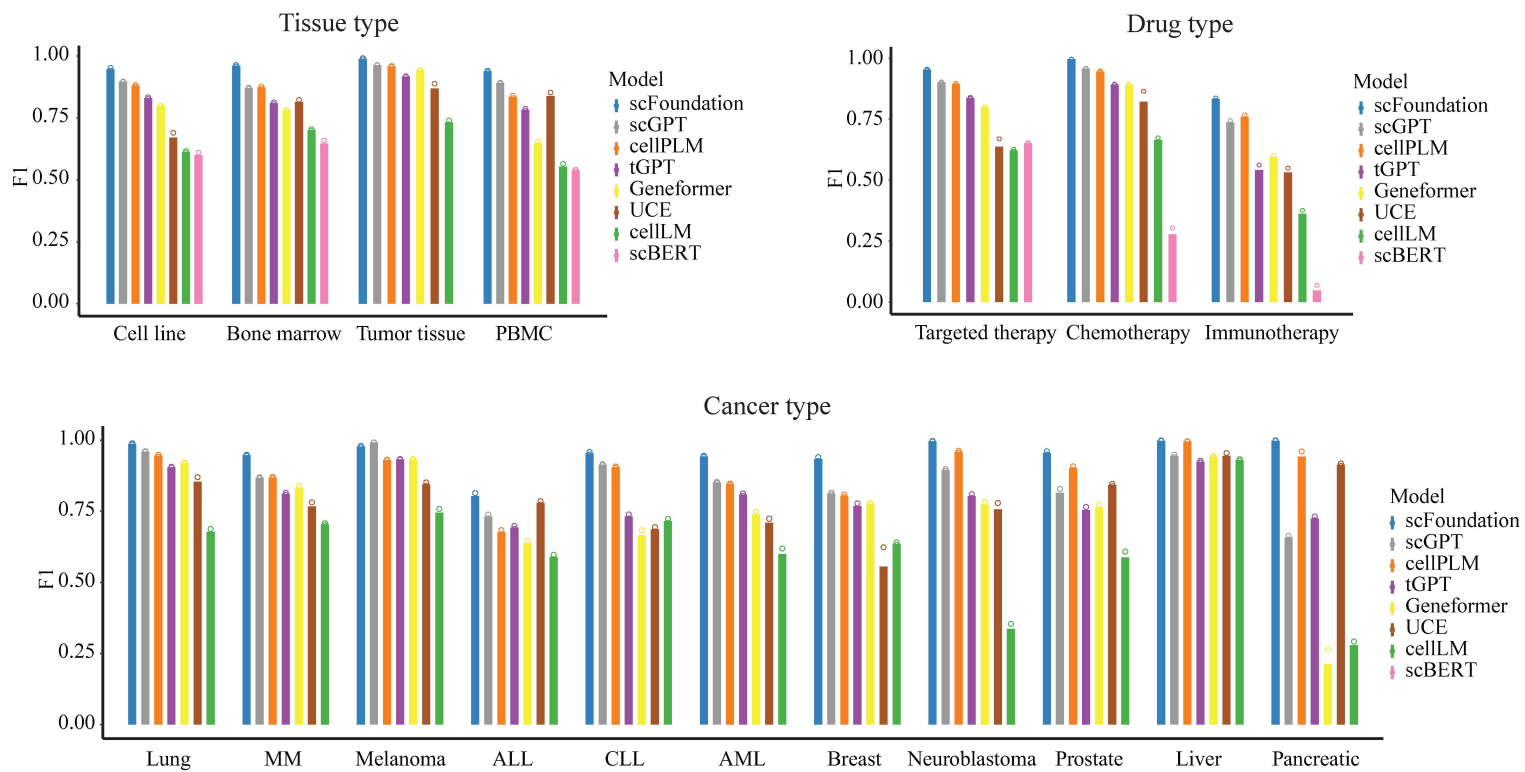

Fig. 3

**a. Results of Layer-freezing in cross-data scenario**

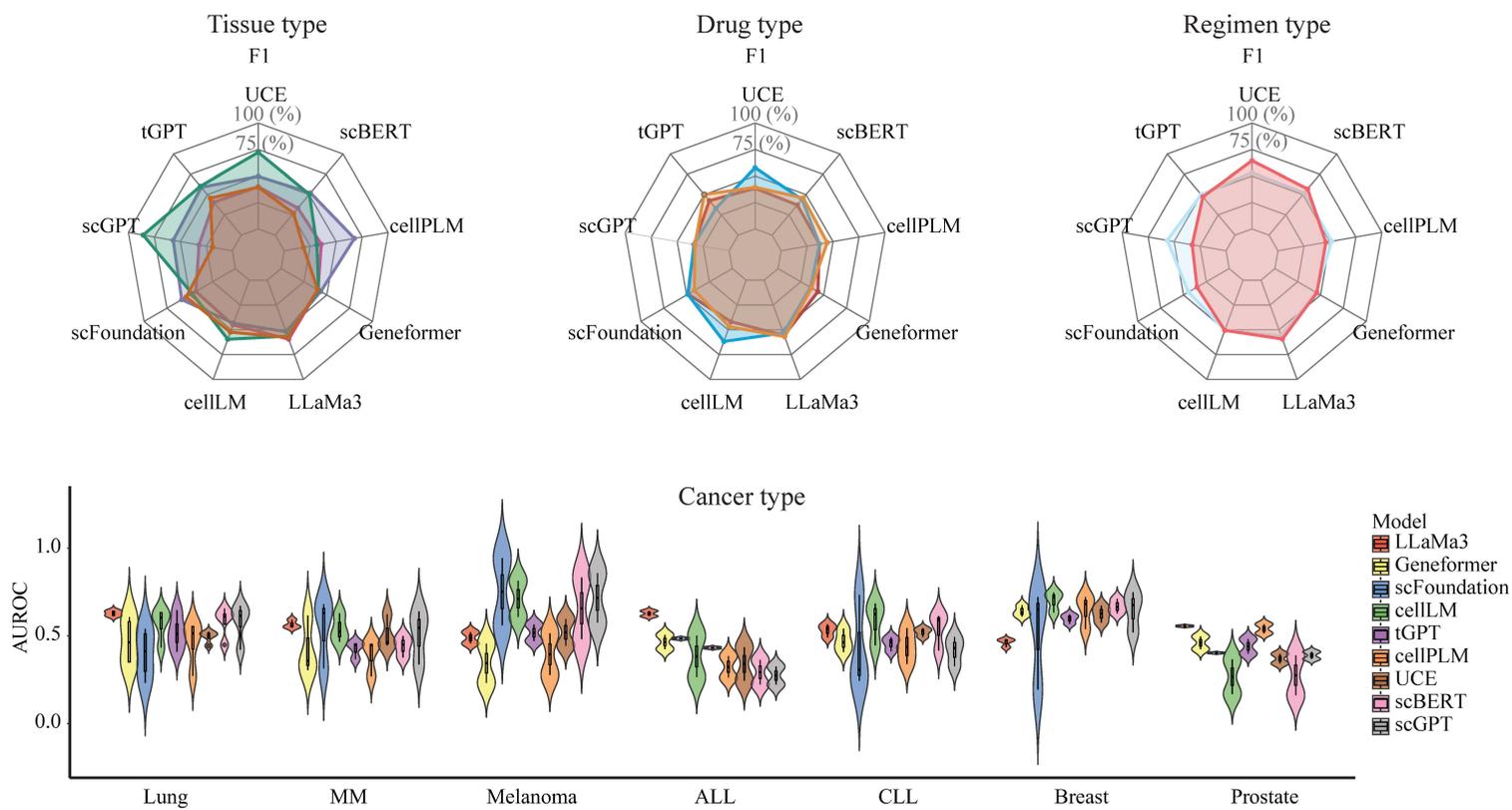

**b. Results of Fune-tuning in cross-data scenario**

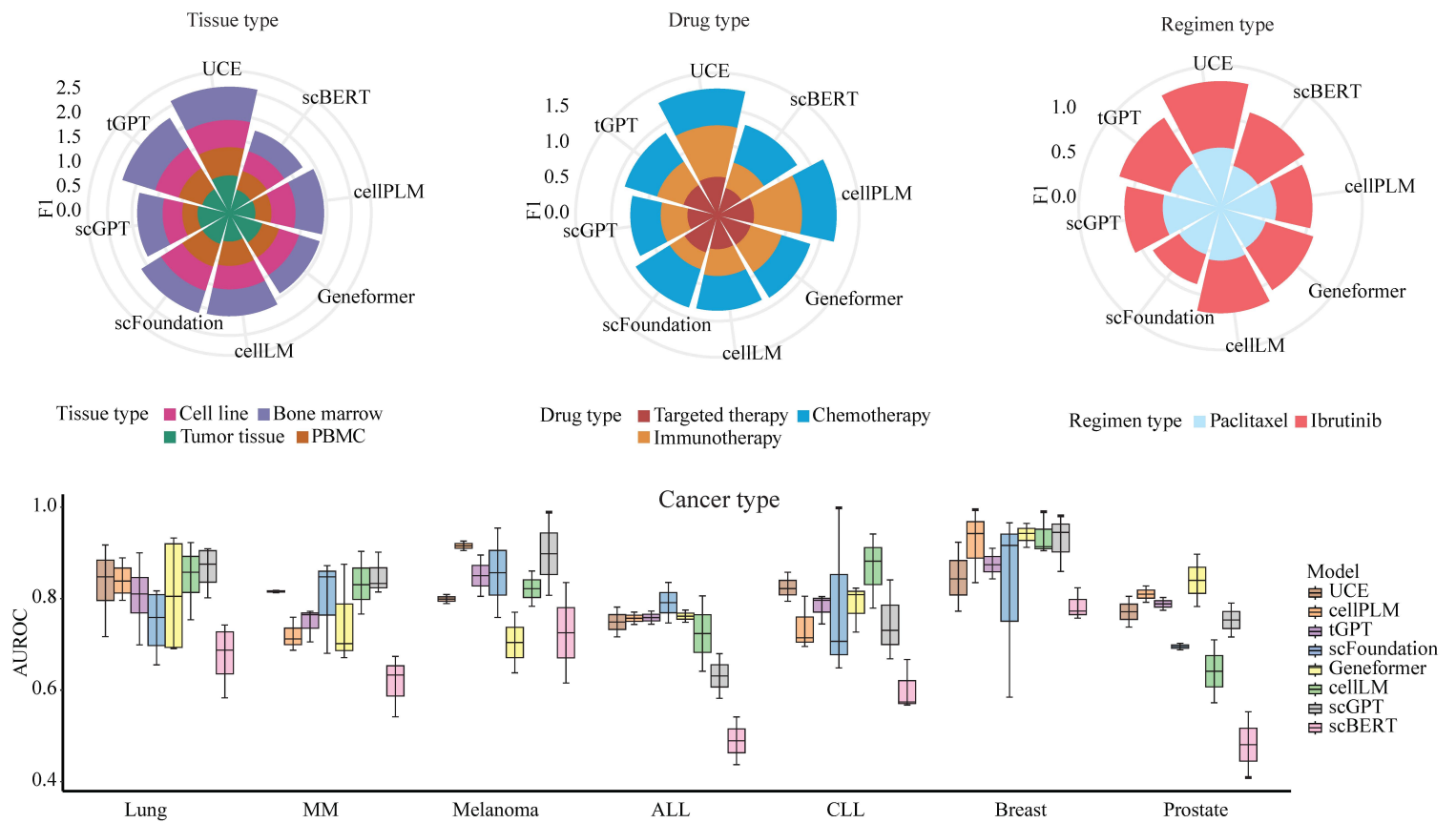

Fig. 4

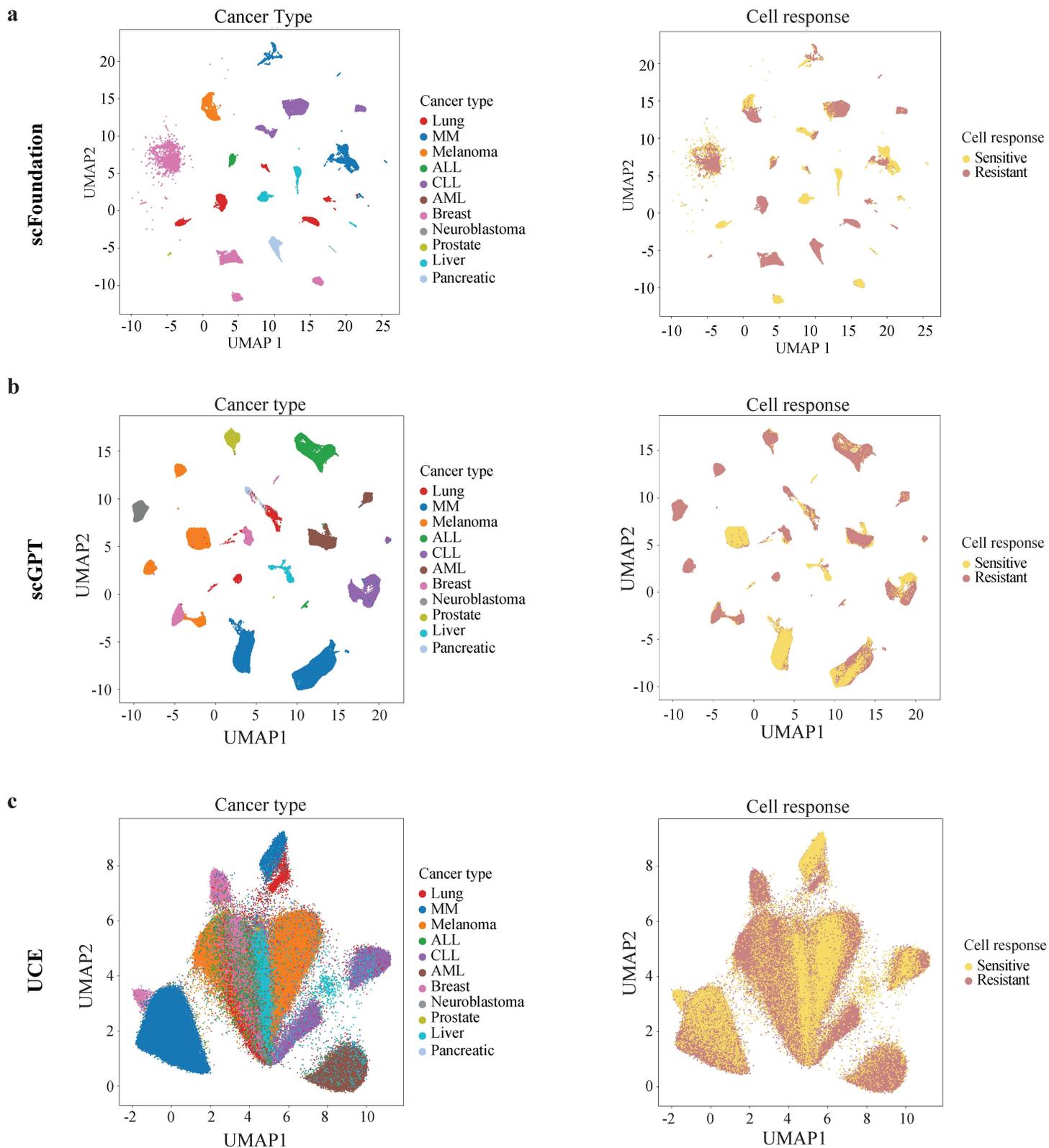

Fig. 5

a

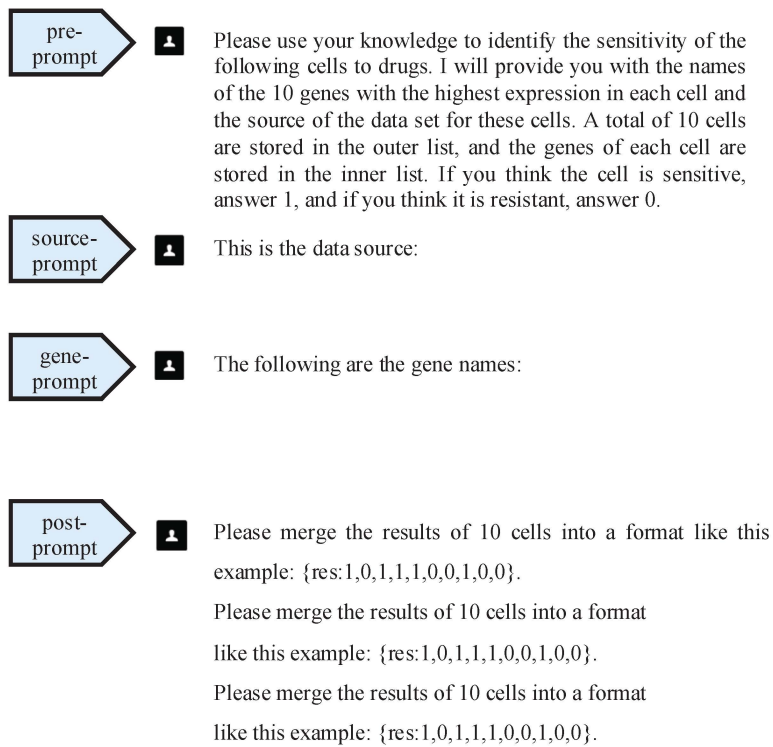

b

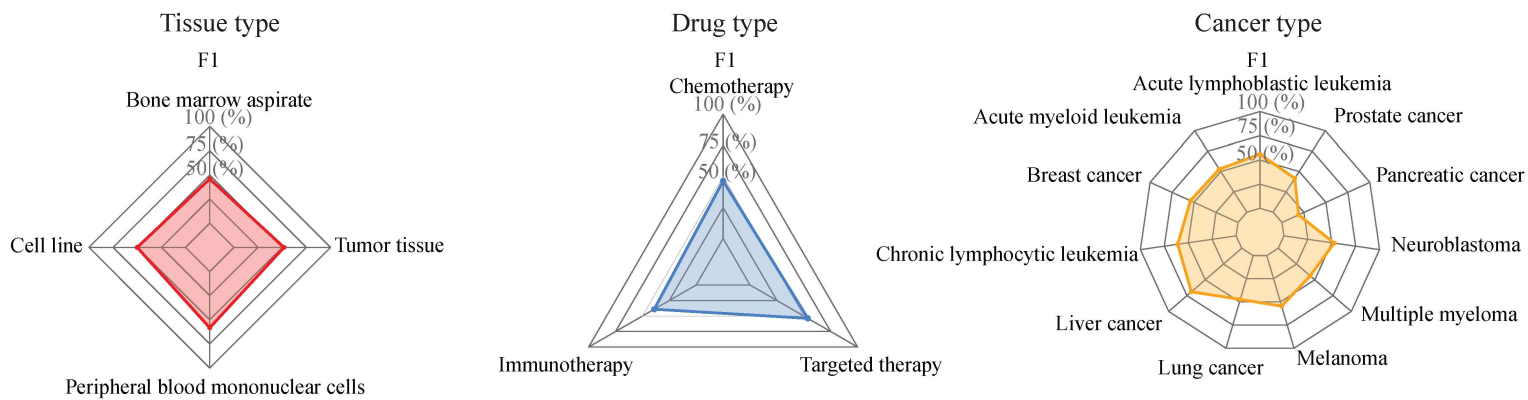

Fig. 6

a

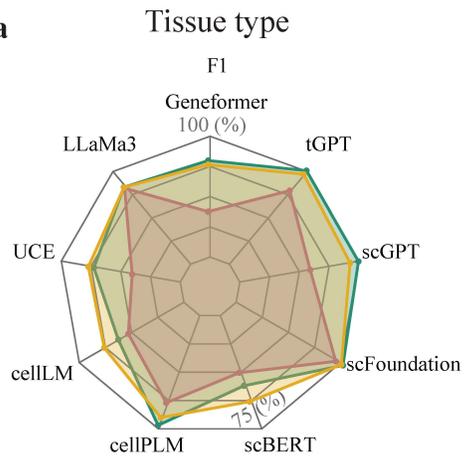
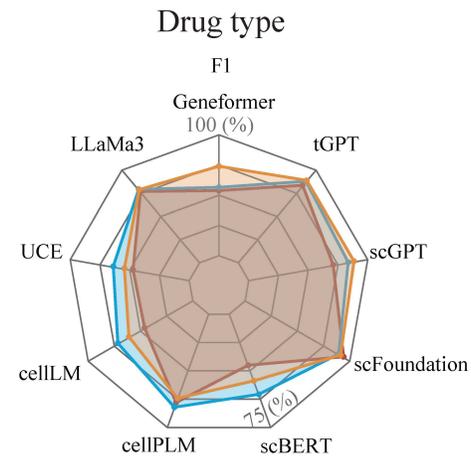
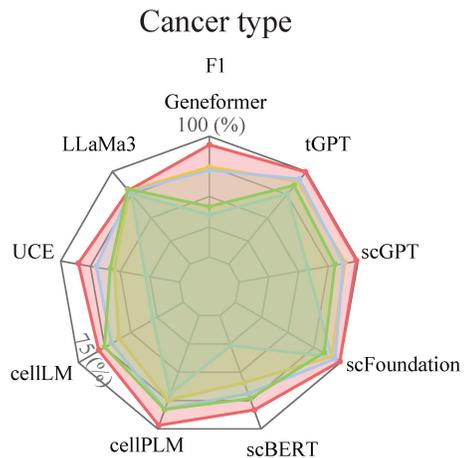

b

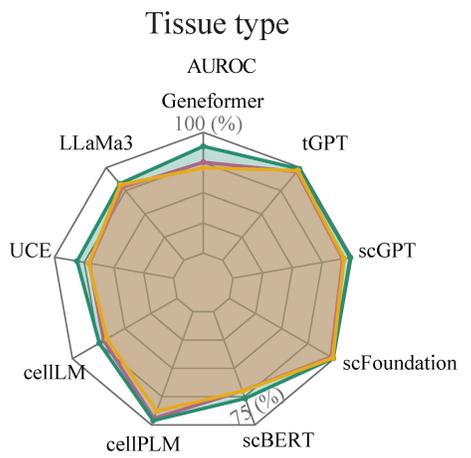
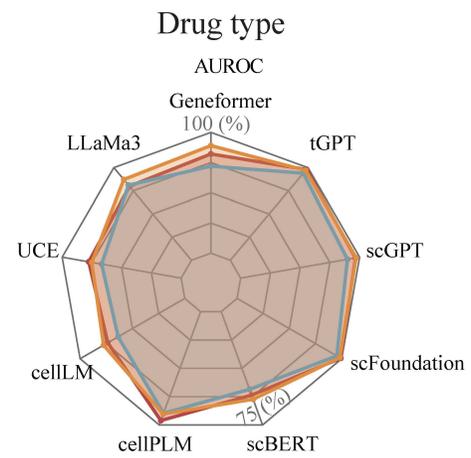
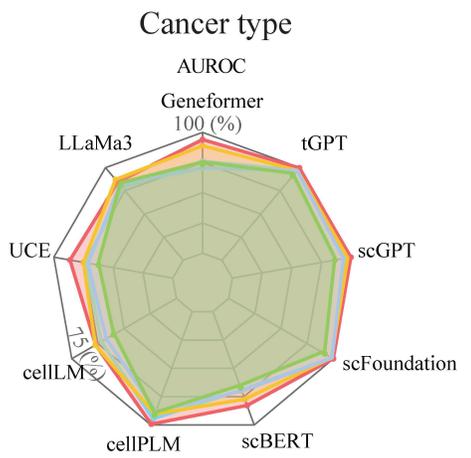

Fig. 7

| | Encoder-decoder? | Input Embedding | scLLM? | Param (m) | Output Dim | Inference speed (it/s) | Inference time (s) | Training speed (it/s) | Training time (s) | Year |
|---|---|---|---|---|---|---|---|---|---|---|
| tGPT | ✗ | Genes | ✓ | 124.5 | 1024 | 9.02 | 1154.56 | 4.4 | 563.2 | Feb-22 |
| scBert | ✗ | Expressions and genes | ✓ | 8.9 | 200 | 14.31 | 42.93 | 6.19 | 18.57 | Sep-22 |
| Geneformer | ✗ | Genes | ✓ | 10.7 | 256 | 1.89 | 378 | 1.3 | 260 | May-23 |
| cellLM | ✗ | Expressions and genes | ✓ | 62.8 | 512 | 3.34 | 13.36 | 1.38 | 5.52 | Jun-23 |
| scFoundation | ✓ | Expressions and genes | ✓ | 121.2 | 3072 | 69.98 | 69.98 | 23.26 | 23.26 | Jun-23 |
| scGPT | ✓ | Expressions and genes | ✓ | 52.5 | 512 | 1.44 | 368.64 | 1.43 | 366.08 | Jul-23 |
| cellPLM | ✓ | Expressions | ✓ | 66.6 | 512 | 51.74 | 51.74 | 29.37 | 29.37 | Oct-23 |
| UCE | ✓ | Expressions | ✓ | 849.9 | 1280 | 1.71 | 102.6 | 1.06 | 63.6 | Nov-23 |
| LLaMa3-8B | ✗ | NLP | ✗ | - | 4096 | - | - | - | - | - |
| GPT4o-mini | ✗ | NLP | ✗ | - | - | - | - | - | - | - |